\documentclass[10pt,twocolumn,letterpaper]{article}

\usepackage[pagenumbers]{cvpr} % To force page numbers, e.g. for an arXiv version
\usepackage{amsmath}
\usepackage{mathtools}
\usepackage[dvipsnames,table]{xcolor}
\newcommand{\method}{DreamHOI}

\usepackage{algorithm} 
\usepackage{algpseudocode}
\usepackage{bm}
\usepackage{bbm}

\definecolor{cvprblue}{rgb}{0.21,0.49,0.74}
\usepackage[pagebackref,breaklinks,colorlinks,citecolor=cvprblue]{hyperref}

\def\paperID{109} % *** Enter the Paper ID here
\def\confName{3DV\xspace}
\def\confYear{2025\xspace}

\newcommand\rurl[1]{%
  \href{https://#1}{\nolinkurl{#1}}%
}

\newcommand{\printfnsymbol}[1]{%
        \textsuperscript{\@fnsymbol{#1}}%
}

\title{{\method}: Subject-Driven Generation of 3D Human-Object Interactions with Diffusion Priors}

\author{Hanwen Zhu\textsuperscript{1,2\dag}
\quad
Ruining Li\textsuperscript{1*}
\quad 
Tomas Jakab\textsuperscript{1*} \\
\textsuperscript{1}University of Oxford \quad
\textsuperscript{2}Carnegie Mellon University \\
{\tt\small thomaszh@cs.cmu.edu \quad \{ruining, tomj\}@robots.ox.ac.uk}\\[0.1em]
\small\rurl{DreamHOI.github.io}
}

\begin{document}
\twocolumn[\maketitle\vspace{-3em}% \begin{figure*}
% \centering
% \includegraphics[width=\linewidth]{figures/teaser.pdf}
% \caption{....}
% \label{fig:teaser}
% \end{figure*}

\begin{center}
    \includegraphics[width=0.99\textwidth]{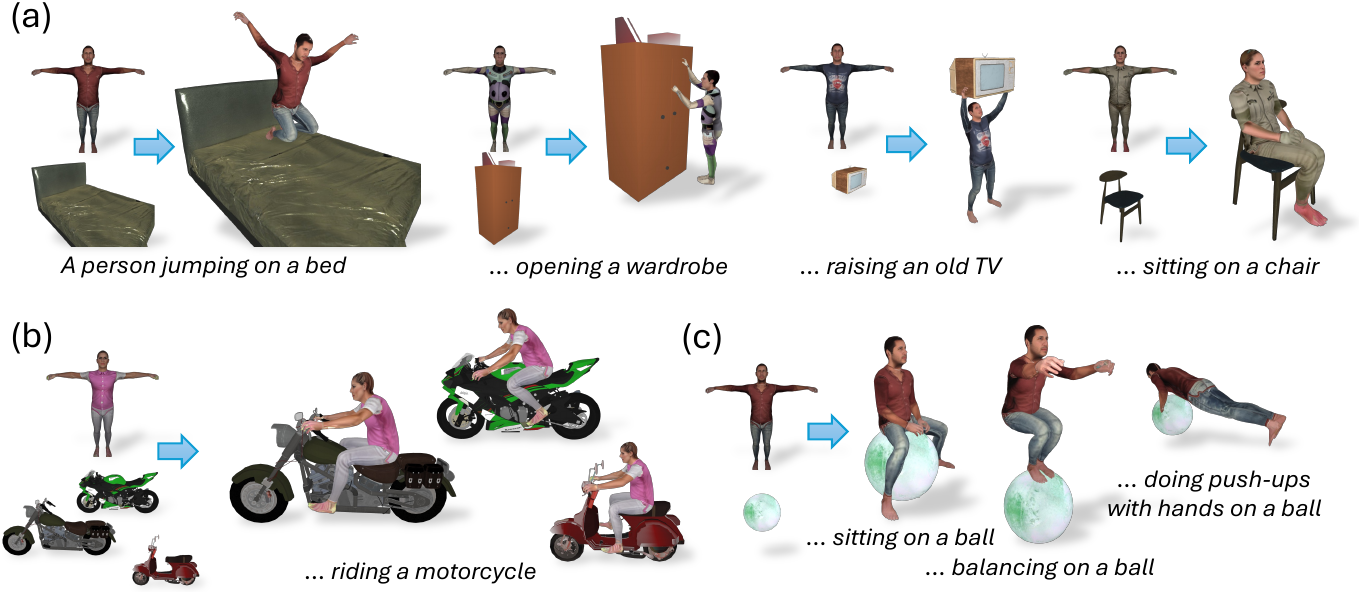}
    \captionof{figure}{
    \textbf{Generated Human-Object Interactions by \method.}
    (a-c) \method~takes as inputs a skinned human model, an object mesh, and a textual description of the intended interaction between them. It then poses the human model to create realistic interactions.
    (b) Given the same interaction description, the generated pose naturally conforms to the intricacies of the input object to be interacted with. 
    (c) Given a fixed object, the generated poses vary faithfully to different intended interactions.
    }
    \label{fig:teaser}
\end{center}
\bigbreak]
\def\thefootnote{\dag}\footnotetext{Work done while at University of Oxford.}\def\thefootnote{\arabic{footnote}}
\def\thefootnote{*}\footnotetext{Equal advising.}\def\thefootnote{\arabic{footnote}}
\begin{abstract}
We present \method, a novel method for zero-shot synthesis of human-object interactions (HOIs), enabling a 3D human model to realistically interact with any given object based on a textual description.
The complexity of this task arises from the diverse categories and geometries of real-world objects, as well as the limited availability of datasets that cover a wide range of HOIs.
To circumvent the need for extensive data, we leverage text-to-image diffusion models trained on billions of image-caption pairs. 
We optimize the articulation of a skinned human mesh using Score Distillation Sampling (SDS) gradients obtained from these models, which predict image-space edits.
However, directly backpropagating image-space gradients into complex articulation parameters is ineffective due to the local nature of such gradients. 
To overcome this, we introduce a dual implicit-explicit representation of a skinned mesh, combining (implicit) neural radiance fields (NeRFs) with (explicit) skeleton-driven mesh articulation.
During optimization, we transition between implicit and explicit forms, grounding the NeRF generation while refining the mesh articulation. 
We validate our approach through extensive experiments, demonstrating its effectiveness in generating realistic HOIs.
The code can be found on the project page at {\small\url{https://DreamHOI.github.io/}}.
\end{abstract}

\section{Introduction}
\label{sec:intro}
The goal of this work is to create a method that can make existing 3D human models realistically interact with any given 3D object, conditioned on a textual description of their interaction.
For instance, given a 3D motorcycle and ``riding'' interaction, we aim to deform the 3D human model so that it realistically appears to ride the given motorcycle (\cref{fig:teaser}).
This task allows for automated population of virtual 3D environments with humans that naturally interact with objects, which has significant implications for industries such as movie and video game production as well as product advertisement.
% This task has a broad range of applications; it enables an automatic population of virtual 3D environments with humans that interact naturally with objects within them.
% This has significant implications for fields such as movie and video game production or product advertisement.

The challenge in achieving open-world synthesis of human-object interaction (HOI) stems from the fact that the target deformation of the human model is influenced not only by the specified interaction but also by the actual geometry of the object.
% This task presents a significant challenge, as the deformation of the human model is influenced not only by the described interaction but also by the actual geometry of the object. 
The riding posture on a cruiser motorcycle typically differs from that on a scooter or sports motorcycle, as shown in Figure \ref{fig:teaser}b; moreover, individual cruiser motorcycles have distinct geometries, each necessitating specific deformations of the character to accommodate them.
To train a model for this task in a \emph{supervised} manner requires triplets of 3D objects, interactions, and target human poses.
While a few such datasets have been recently collected \cite{bhatnagar22behave, jiang2023chairs, zhang2022couch, taheri2020grab, xu2021d3d, li2023object}, methods trained on them \cite{peng2023hoi,wu2024thor,diller2024cg} are constrained by the limited object and interaction coverage in these datasets.
Constructing even larger datasets of immensely diverse real-world objects which humans can interact with is exceptionally difficult and expensive.
% There is an immense variety of objects in the world with which humans can interact, each exhibiting endless variations in geometry, and there are countless ways in which humans can engage with them.
% -> put in limitations \ray{Should we also mention only human justifies for large-scale data collection, while our method can be extended to other categories relatively easily?}

% Since each motorbike can have a different geometry, the deformed character must respect that. 
% For instance, a cruiser is ridden in a different pose than a scooter or a sports bike. 
% We aim to automatically infer the appropriate pose from the object's geometry.

% This task presents a significant challenge due to the scarcity of data. 
% Although there has been some recent success in human motion generation [TODO: reference], the study of human-object interaction has been confined to the context of human pose prediction rather than generation. 
% Furthermore, it has been limited to a small selection of objects [TODO: reference the paper/dataset called CHAIRS]. 
% The difficulty arises from the countless objects that exist, each with infinite variations in geometry, and the myriad ways in which humans can interact with them. 
% This complexity far exceeds that of human motion generation. 
% Moreover, compiling a sufficiently large dataset that encompasses pairs of objects, interactions, and human 3D poses is an exceedingly difficult and costly endeavor.

To bypass tedious data collection, we propose to distill the guidance for HOI synthesis from off-the-shelf text-to-image diffusion models that have been trained on large-scale image-caption pairs.
% Given an object mesh and a (skinned) human mesh, together with a textual description of their interaction, our goal is to optimize the articulation of the human mesh.
These models can act as a ``critic'' and provide image-space gradients through Score Distillation Sampling (SDS)~\cite{poole2022dreamfusion}.
These gradients indicate how to modify the image to better align with the given text prompt and can be used to optimize the actual image parameters through backpropagation.
This approach has proven effective in text-to-3D generation, where the underlying 3D representation, such as NeRF~\cite{mildenhall2020nerf}, is rendered differentiably to obtain images for a text-to-2D diffusion model to ``critique''. The local ``edits'' proposed by the 2D model are then backpropagated into the 3D representation space to optimize it~\cite{poole2022dreamfusion,lin2023magic3d,shi23mvdream, wang2023score, melas2023realfusion, jakab2024farm3d}.
However, applying this approach to our problem is not straightforward.

In our case, the 3D scene consists of an input object mesh and a skinned human mesh. 
The goal is to pose the human, \ie, to optimize the articulation parameters of the skinned human mesh.
In theory, we can differentiably render the two meshes together and apply SDS. 
However, this does not work well in practice (\cref{sec:comparisons}) due to the \emph{local} nature of the image-space SDS gradients.
Intuitively, the image-space gradients are \emph{local} edits, indicating where in the image something should be added, removed, or slightly modified, which makes it challenging to directly derive the \emph{structural} changes required for the skeleton.
Moreover, gradient-based optimization is prone to local optima.
Updating the skeleton to its optimal location may necessitate human poses less aligned with the textual prompt during intermediate steps.
Consequently, directly optimizing the articulation parameters using SDS often leads to convergence in sub-optimal poses.

To overcome this challenge, we draw inspiration from \cite{jakab2020self}, which translates between an implicit pixel representation and an explicit skeleton-based human pose representation in 2D, and introduce a dual implicit-explicit 3D representation of the skinned human mesh.
The implicit component is represented by a neural radiance field (NeRF)~\cite{mildenhall2020nerf}, while the explicit component comprises the input skinned mesh along with its articulation parameters, \ie, bone rotations.
To convert the implicit representation into the explicit one, we employ a regressor that predicts the bone rotations of the skinned mesh from multi-view renderings of the NeRF. 
Conversely, to revert to the implicit representation, we initialize the NeRF using the posed mesh. 
The implicit representation (\ie, NeRF) then facilitates distillation from pre-trained diffusion models.
However, it introduces a significant challenge in maintaining the identity of the articulated character during optimization. 
To address this issue, we periodically transition between the implicit and explicit representations. 
Specifically, after a number of optimization steps on the NeRF, we convert it to the explicit representation. 
By doing so, we ensure that the character's identity is preserved, as we simply substitute the bone rotations of the skinned human with those predicted by the regressor, without changing the body shape or appearance.
Subsequently, we re-initialize the NeRF using the posed human mesh and continue the optimization process.

We also make an important observation: while multi-view text-to-image diffusion models such as \cite{shi23mvdream} are better at generating 3D assets than their single-view counterparts, we find that they tend to have an insufficient understanding of human-object interactions, potentially due to their extended fine-tuning on synthetic renderings.
Therefore, we propose a simple and effective technique to combine the guidance from a multi-view text-to-image diffusion model with that from a state-of-the-art single-view model \cite{deepfloyd2023if}, leveraging their respective strengths.

To summarize, our contributions are:
\begin{enumerate}
\item We introduce a novel \emph{zero-shot} approach dubbed \method{} for open-world synthesis of human-object interactions with a given human mesh, object mesh, and textual description of the interaction between them.
\item We propose a dual implicit-explicit representation of a skinned human mesh that allows us to harness the power of large text-to-image diffusion models to optimize its articulation parameters.
% (3) We propose a simple and effective mixing strategy of multiview and single-view diffusion models.
\item We demonstrate the effectiveness of our approach with extensive qualitative and quantitative experiments.
\end{enumerate}

\section{Related Work}
\label{sec:related}

\paragraph{3D Generation.}
% Unsup3D, dove, magicpony, farm3d, 3dfauna

% DreamFusion, Magic3D

% Zero123, MVDream, IM-3D, SV3D, CAT3D

% Composition 3D generation: 
Early attempts on 3D generation~\cite{wang2018pixel2mesh, sitzmann2019deepvoxels, wu2020unsup3d, yu2021pixelnerf, wu2023dove, wu2023magicpony, jakab2024farm3d, li2024learning, szymanowicz2024splatter} formulate the task as single/few-view reconstruction, and propose learning frameworks that regress the 3D model with various representations from the conditioning image(s).
However, these models are category-specific and do \emph{not} allow open-domain generation.
Recent works have explored open-domain 3D generation using pre-trained image/video generators.
Such generators, powered by diffusion models~\cite{ho2020denoising, rombach2022stablediffusion, deepfloyd2023if}, are expected to acquire an implicit 3D understanding through their pre-training on Internet-scale images/videos.
DreamFusion~\cite{poole2022dreamfusion} and follow-ups~\cite{lin2023magic3d, wang2023score, melas2023realfusion, wang2023prolificdreamer, tang2024dreamgaussian}
distill 2D diffusion models to extract 3D models from text or image prompts.
To improve the generation process, authors have proposed to fine-tune pre-trained generators on datasets of multi-view images~\cite{liu23zero-1-to-3, shi23mvdream, zheng2023free3D, li2024instant3d, melas-kyriazi2024IM3d, voleti2024sv3d, gao2024cat3d}.
However, since these datasets consist mostly of scenes with single objects, the fine-tuned models often struggle with generating compositions and interactions between objects.
In this work, we build upon both single-view and multi-view diffusion models to enable the generation of human-object interactions (HOIs) in 3D.

\paragraph{Compositional Generation.}
Prior work~\cite{liu2022compositional} has shown that existing image generators tend to struggle with composing multiple concepts into a single image.
While efforts have been made to enhance compositionality in image generative models~\cite{du2020compositional, liu2022compositional}, such techniques are not easily portable to distillation and hence 3D generation.
Recent works instead extend the distillation approach to compositional 3D generation.~\citet{po2024compositional} introduce locally conditioned distillation, masking the 2D diffusion model's prediction according to user-specified bounding boxes to generate composition.
CG3D~\cite{vilesov2023cg3d} optimizes individual objects and their relative rotation and translation via score distillation sampling (SDS,~\cite{poole2022dreamfusion}), subject to additional physical constraints.
Similarly, SceneWiz3D~\cite{zhang2023scenewiz3d} proposes a hybrid representation to achieve scene generation, optimizing individual objects via SDS and a scene configuration.
GALA3D~\cite{zhou2024gala3d} leverages a large language model to extract scene layout, which is then used to further refine the generated scene through layout-conditioned diffusion.
ComboVerse~\cite{chen2024comboverse} approaches compositional image-to-3D generation by leveraging a pre-trained single-view 3D reconstructor to generate the individual objects, followed by spatial relationship optimization using SDS.
Concurrent to our work is InterFusion~\cite{dai2024interfusion}, which synthesizes human-object interactions based on textual prompts.
This method selects an anchor pose from a codebook of poses in response to a textual prompt.
Subsequently, it synthesizes one neural radiance field (NeRF) representing a human body and another for an object, both designed to fit the anchor pose.
However, these two NeRFs are optimized from scratch, offering no control over the identities of the human and the object.
In contrast, our human-object interaction (HOI) generation pipeline takes existing 3D subjects (\ie, the object mesh and the human mesh) as inputs and synthesizes the interaction with appropriate pose deformation.
Our approach not only affords greater user control but is also more practical as it can be easily integrated into existing pipelines for the production of virtual 3D environments.
% It trains a feed-forward network to predict human poses directly from the input prompts, and jointly optimizes \emph{two} 3D models for the human and the object respectively using both local and global SDS. Different from these methods where users \emph{cannot} precisely control the individual objects' identity, our HOI generation pipeline takes the 3D subjects (\ie, the object mesh and the human mesh) as input, and synthesizes the interaction with proper pose deformation, providing users with extra controllability.

\paragraph{Data-Driven Human-object Interaction Synthesis.}
Since digital humans play an essential role in numerous applications ranging from AR/VR to gaming and movie production, prior works have considered \emph{training} generative models of static human pose or dynamic human motion conditioned on an object or environment using curated HOI datasets, enabling HOI synthesis at test time.
Earlier works~\cite{zhao2022compositional, zhang2020generating, hassan2021populating} use a combination of affordance prediction and semantic heuristics to synthesize a static human pose. More recent works~\cite{wu2024thor, peng2023hoi, diller2024cg} learn human motion diffusion models~\cite{tevet2022human} to generate a dynamic motion clip, conditioned on the encoded object or scene.
The key difference between these training-based methods and our work is that our method extracts the HOI solely from a pre-trained diffusion model and does \emph{not} require bespoke training data for this task.
Consequently, our method can be easily extended to other categories by switching to a different deformation model, \eg, using SMAL~\cite{zuffi173d-menagerie} for quadruped animals~\cite{zuffi173d-menagerie}, whose motion data is not available at scale.

\paragraph{Human Deformation Models.}
To accurately reason about object pose and motion in images and videos,
authors have developed deformation models capable of capturing the shape deformation space using low-dimensional latent codes~\cite{loper15smpl, zuffi173d-menagerie, li2024learning, zuffi2024varen} or conditioning inputs~\cite{sumner2007embedded, li2024dragapart, li2024puppetmaster}.
The most popular model for human bodies is SMPL~\cite{loper15smpl},
which decomposes the 3D body into shape-dependent deformations,
statistically learned from a large-scale 3D body scan dataset,
and pose-dependent deformations, represented by joint rotations which are then used to deform body vertices with linear blend skinning~\cite{chadwick1989layered, magnenat1988abstract}.
Numerous follow-ups have sought to improve this model. For instance, MANO~\cite{romero2017mano}, or SMPL+H, extends SMPL to also fit human hands. SMPL-X~\cite{pavlakos2019smplx} goes one step beyond, 
facilitating the 3D model to capture both fully articulated hands and an expressive face.
STAR~\cite{osman2020star} leverages a more compact shape representation and a larger human scan dataset than SMPL,
which mitigates over-fitting and helps the model generalize better to new bodies.
Meanwhile, orthogonal efforts~\cite{cao2017_openpose, wei2016_openpose_cpm, simon2017_openpose_hand, ye2023decoupling, wang2024tram} have focused on training pose estimators which regress shape parameters of these body models from in-the-wild images and videos. In this work, we explore the possibilities of using these human body models and their estimators to allow high-quality HOI generation.

\section{Method}
\label{sec:methods}

\begin{figure*}
\centering
\includegraphics[width=\linewidth]{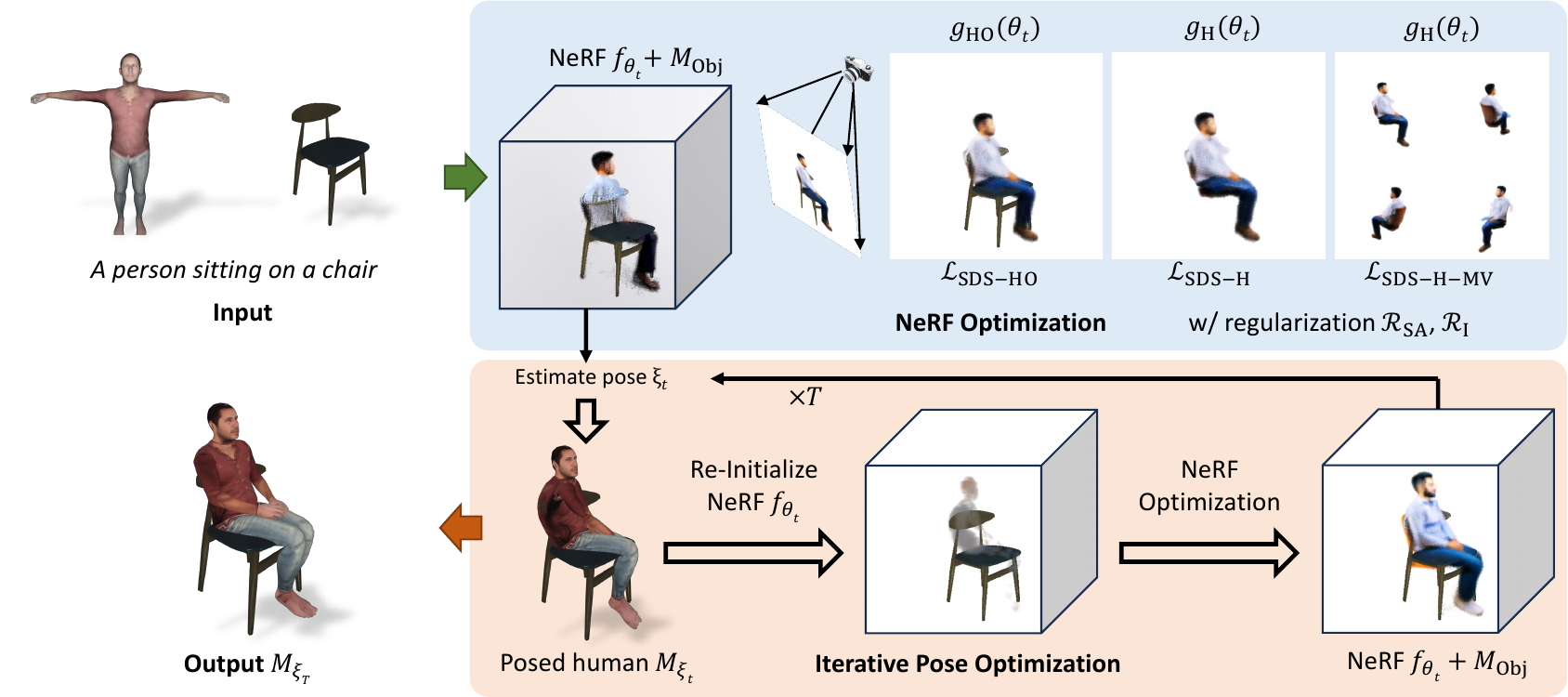}
\caption{\textbf{Overview of \method}.
Our method takes a human identity (in the form of a skinned body mesh) and an object mesh $M_{\text{Obj}}$ (\eg, a 3D chair), together with their intended interaction (as a textual prompt, \eg, ``sit''), as input.
It first fits a NeRF $f_{\theta_0}$ for the human using a mixture of diffusion guidance and regularizers, and then estimates its pose $\xi_0$.
The posed human mesh $M_{\xi_t}$ is used to re-initialize and further optimize the NeRF $f_{\theta_{t}}$, for iterations $t\leq T$.
The final output is the posed human $M_{\xi_T}$ at the last iteration. See~\cref{sec:methods}.
}
\label{fig:method}
\end{figure*}

Given an object mesh $M_{\text{Obj}}$ (\eg, a 3D ball) and an intended interaction $r$ as a textual prompt (\eg, ``sit''), our goal is to automatically optimize the pose parameters $\xi$ of a skinned human model $M_\xi$.
The pose should realistically reflect the interaction $r$ with the object (\ie, the character $M_\xi$ is sitting on the ball $M_{\text{Obj}}$) (\cref{fig:teaser}).
To achieve this, we propose a dual implicit-explicit representation of the skinned human model (\cref{sec:implicit-explicit}), which enables us to leverage the image-space gradients from text-to-image diffusion models to optimize the 3D human's pose parameters.
We compose the human-object scene using the dual representation (\cref{sec:human-object-scene}) and apply a mixture of SDS guidance from both multi-view and single-view diffusion models (\cref{sec:guidance-mixture}).
We introduce additional regularizers to ensure the appropriate size of the human model and the consistency of the rendered scene (\cref{sec:regularizers}).
Finally, we integrate all components and devise an iterative optimization procedure for the pose parameters $\xi$ (\cref{sec:optimization}).
See \cref{fig:method} for an overview.

\subsection{Preliminaries}
\label{sec:prelim}
\paragraph{Neural Radience Fields (NeRFs).}
A neural radiance field (NeRF~\cite{mildenhall2020nerf}) represents a 3D scene as a function parametrized by $\theta$:
\[
f_{\theta} : \bm\mu\mapsto(\bm c,\tau).
\]
This function maps a 3D location $\bm\mu=(x,y,z)$ to a color $\bm c=(r,g,b)$ and a volume density $\tau\ge0$\footnote{In this work, for simplicity, the RGB color is \emph{not} view-dependent.}.
NeRFs can be rendered differentiably by aggregating the color of corresponding locations,
and hence optimized via gradient descent using posed multi-view images.
% Specifically, for each pixel $\bm u$, we sample $K$ points $\left\{ \bm \mu_k \right\}_{k=1}^K$
% along the ray cast from the camera $\bm \mu_c$ to the pixel $\bm u$,
% and obtain the pixel's color $\bm c_{\bm u}$ as the weighted average of the color $\bm c_k$ of each point $\bm \mu_k$:
% \[
% \bm c_{\bm u} = \sum_k w_k \bm c_k,
% \]
% where $(\bm c_k, \tau_k) = f_\theta(\bm \mu_k)$.
% The weights $w_k$ are designed to
% prioritize points with larger density $\tau_k$, as they contribute more significantly to the final color. Conversely, points that are more occluded by other points closer to the camera contribute less
% (\ie, have a smaller $w_k$).

% A Neural Radiance Field (NeRF)~\cite{mildenhall2020nerfrepresentingscenesneural} is a function parameterized by $\theta$ that represents a 3D scene:
% \[
% f_{\theta} : (x,y,z)\mapsto(\bm c,\tau)
% \]
Specifically, for each pixel $\bm u$, we sample $N$ points $\left\{ \bm \mu_i \right\}_{i=1}^N$ along the ray cast from the camera $\bm \mu_c$ to the pixel $\bm u$.
We then obtain $(\bm c_i, \tau_i) = f_{\theta}(\bm\mu_i)$, sorted by distance to the camera $d_i=\Vert\bm\mu_i-\bm\mu_{c}\Vert$.
Finally, we obtain the color $\bm c$ of the pixel by alpha-compositing along the ray, as described in~\cite{mildenhall2020nerf}:
\begin{gather}
\bm c=\sum_iw_i\bm c_i ,\quad w_i=\alpha_i\prod_{j<i}(1-\alpha_j)
,\\ \alpha_i=1-\exp(-\tau_i\Vert\bm\mu_i-\bm\mu_{i+1}\Vert)
\label{alpha}
.\end{gather}

% and obtain $(\bm c_i, \tau_i) = f_w(\bm\mu_i)$
%  % We then obtain  = f_{\theta}(\bm\mu_i)$, sorted by distance to the camera $d_i=\Vert\bm\mu_i-\bm\mu_{c}\Vert$.
% Finally, we obtain the color $\bm c$ of the pixel by alpha-compositing along the ray, as described in~\cite{mildenhall2020nerfrepresentingscenesneural}:
% \begin{gather}
% \bm c=\sum_iw_i\bm c_i ,\quad w_i=\alpha_i\prod_{j<i}(1-\alpha_j)
% ,\\ \alpha_i=1-\exp(-\tau_i\Vert\bm\mu_i-\bm\mu_{i+1}\Vert)
% \label{alpha}
% .\end{gather}

% \paragraph{Rendering NeRFs and Meshes.}
% We hence differentiably obtain the rendered image $\bm x_{\text{G}}$ with both the NeRF and the mesh $M_{\text{Obj}}$.

\paragraph{Score Distillation Sampling (SDS).}
Score Distillation Sampling (SDS~\cite{poole2022dreamfusion}) leverages pre-trained text-to-image diffusion models such as Stable Diffusion~\cite{rombach2022stablediffusion} to align a NeRF with a textual prompt.
Assuming the learned denoiser $\hat\epsilon(\bm{x}_t; y, t)$ which predicts the sampled noise $\epsilon$ given the noisy 2D image $\bm{x}_t$, noise level $t$ and textual prompt $y$,
SDS computes the gradient: 
\begin{equation} \label{eq:sds-loss}
\nabla_\theta\mathcal{L}_{\text{SDS}}(\bm{x}) =
\mathbb{E}_{t,\epsilon}\left[
w(t)(\hat\epsilon(\bm{x}_t; y, t)-\epsilon)\frac{\partial \bm{x}}{\partial \theta}
\right]
\end{equation}
with respect to the NeRF parameters $\theta$.
Here, the clean image $\bm{x}$ is obtained via NeRF rendering,
and $w(t)$ is a weighting function.
DreamFusion~\cite{poole2022dreamfusion} achieves text-to-3D generation by iteratively updating $\theta$ using~\cref{eq:sds-loss} from a randomly initialized NeRF.

% and the time step $t$ is sampled from a distribution as described in \ref{todo}. % $t\sim\mathcal{U}(0.02,0.98)$ for numerical stability as in~\cite{poole2022dreamfusion}.

\paragraph{Skinned 3D models.}
A skinned 3D mesh $M_\xi = (V\in \mathbb{R}^{|V|\times 3}, E)$ is parametrized by $B$ bone rotations $\xi = \left\{\xi_b\right\}_{b=1}^B \in SO(3)$.
To deform $|V|$ vertices driven by $B$ bones, we define rest-pose joint locations $\mathbf{J} \in \mathbb{R}^{B\times 3}$, a kinematic tree structure $\pi$ defining a parent $\pi(b)$ for each bone $b$, and skinning weights $W \in \mathbb{R}^{|V|\times B}$.
The vertices are then posed by the linear blend skinning equation:
\begin{equation*}\label{e:skinning}
\begin{aligned}
    \hat{V}_i^\text{posed}(\xi) =\left( \sum_{b=1}^B W_{ib} G_b(\xi) G_b(\xi^*)^{-1} \right) \hat{V}_{i}, \\
    G_\text{root} = g_\text{root},
    ~~
    G_b = g_b \circ G_{\pi(b)},
    ~~
    g_b(\xi) = \begin{bmatrix}
        R_{\xi_b} & \mathbf{J}_{b} \\ 0 & 1 \\
    \end{bmatrix},
\end{aligned}
\end{equation*}
where $\xi^*$ denotes the bone rotations at the rest pose\footnote{The $\hat{\cdot}$ indicates $\hat{V}_i$ and $\hat{V}_i^\text{posed}$ are in homogeneous coordinates.}.
In practice, $\mathbf{J}$ and $W$ are either hand-crafted or learned.
In this work, we adopt SMPL+H~\cite{romero2017mano} as the skinned model for human body.
% \thomas{for brevity, we also include the global rotation/translation/scale in $\xi$}
% \tom{this can be done only through the root bone, but I think this is a detail we can omit}

\subsection{Implicit-Explicit Skinned Mesh Representation}
\label{sec:implicit-explicit}
Given a skinned human mesh $M_\xi$, where $\xi$ denotes bone rotations, our goal is to optimize these parameters so that the resulting posed mesh conforms to the desired interaction with a given object.
As demonstrated in \cref{sec:comparisons} and discussed in \cref{sec:supmat-discussions}, optimizing the pose parameters $\xi$ directly by backpropagating image-space SDS gradients is not feasible due to the uninformative nature of these gradients.
This issue is particularly pronounced for complex poses, such as those of humans.
To address this challenge, we propose a dual implicit-explicit representation of a skinned mesh.
The implicit representation consists of a NeRF and serves as a proxy to the skinned mesh, which can be efficiently optimized using image-space SDS gradients.
In the following, we detail the process of translating from the implicit NeRF representation to the posed mesh and back.

\paragraph{Translation from NeRF to Posed Mesh.}
Given a NeRF $f_\theta$ that represents a human, we render a set of images $\bm x_i$ of $f_{\theta}$ from multiple camera viewpoints $c_i$.
Using these multiview images $\bm x_i$, we employ a pose estimator~\cite{zheng2020pamir} to estimate the bone rotations $\xi$ of the skinned mesh $M_\xi$.
We note that our framework can, in principle, be applied to other categories, provided that a pose estimator for the skinned mesh is available, such as in the case of dogs~\cite{Ruegg2023BARC,zuffi173d-menagerie}.

% We note that our framework \emph{is not limited to predicting SMPL parameters}, and in principle supports any parametrized body modeling that can be predicted from (renderings of) 3D objects and converted to a mesh or SDF. For example, our framework in principle also estimating SMPL~\cite{loper15smpl}, SMPL+H (SMPL with hand modeling)~\cite{romero2017mano}, SMPL-X (SMPL with hand and face expressions)~\cite{pavlakos2019smplx}, \todo{animals, dogs}. In particular we use SMPL+H instead of SMPL in our experiments, because the human hand modeling enables more realistic human models in certain semantic relations (\eg ``holding on a handle'').

\paragraph{Translation from a Posed Mesh to NeRF.}
Given a posed mesh $M_\xi$, we construct a NeRF $f_\theta$ by fitting $f_\theta$ on $1$k randomly sampled points $\bm\mu\sim\mathcal{N}(0,I_3)$.
Recall that $f_\theta(\bm\mu)=(\bm c, \tau)$ maps a 3D location $\bm \mu$ to its color $\bm c$ and volume density $\tau\geq 0$.
Specifically, the target color at each sampled point $\bm\mu$ is the color of the mesh vertex of $M_\xi$ that is closest to $\bm\mu$; the target density is $\infty$ is $\bm\mu$ is inside $M_\xi$ and $0$ if outside.
% Specifically, we optimize $\theta$ as a NeRF initialization step such that the density estimate $\tau$ is close to $\infty$ if $\bm\mu$ is inside mesh $M_\xi$, and $0$ if outside; and $\bm c$ is close to the color of the mesh vertex of $M_\xi$ closest to $\bm\mu$.
% given by a fixed human texture (\eg~\cite{smpl_texture_github,casas2023smplitex}).
We optimize $\theta$ in a supervised manner for $10$k iterations, taking about a minute.
% This optimization (for $10$k iterations) takes about a minute.

\subsection{Human-Object Scene Representation}
\label{sec:human-object-scene}
The scene consists of the object mesh $M_{\text{Obj}}$ and the dual implicit-explicit skinned mesh representation of the human which can take the form of either a skinned mesh $M_\xi$ or a NeRF $f_\theta$, with the NeRF being used during optimization.
To render the human-object scene $\bm x_{\text{HO}}=g_{\text{HO}}(\theta)$ consisting of the mesh $M_{\text{Obj}}$ and the NeRF $f_\theta$ we extend the standard volumetric rendering described in Section~\ref{sec:prelim}, which we denote as $g_{\text{HO}}$, to accommodate joint rendering with the mesh.
% The above formulation can be extended for rendering scenes that have parts represented by a mixture of NeRFs and 3D mesh representations.

Recall that to render a pixel $\bm u$, we sample $N$ points $\left\{ 
\bm\mu_i \right\}_{i=1}^N$ along the ray and obtain $P = \left\{(\bm c_i, \tau_i)\right\}_{i=1}^N$ by evaluating $f_\theta$.
In the case where the ray intersects the object mesh $M_{\text{Obj}}$, to account for the effect of $M_{\text{Obj}}$ on pixel $\bm u$, we insert an additional value $(\bm c, \tau=\infty)$ into $P$ at the (first) intersection point $\bm \mu$, where $\bm c$ is the color of $M_\text{Obj}$ at $\bm \mu$.
The density value $\infty$ reflects the opaque nature of $M_\text{Obj}$.
We then apply the same alpha compositing method as in~\cref{alpha}.

% Given the object mesh $M_{\text{Obj}}$, we determine whether each ray intersects $M_{\text{Obj}}$ at a point $\bm\mu_i$.
% If it does, we calculate the distance to the camera $d_i=\Vert\bm\mu_{i}-\bm\mu_c\Vert$ using traditional rendering methods.
% Next, we obtain the RGBA value $\bm c_i,\tau_i$ of $M_{\text{Obj}}$ at point $\bm\mu_i$. 
% -> \ray{isn't $\tau_i$ always 1 if the ray intersects the object?} -> mostly yes, but when the mesh has transparent parts (motorcycle windshield) \tau can be less. this is not what I do in practice for simplicity but I don't think we need to explain
% These values are then inserted into the sorted list $(\bm c_i,\tau_i, \bm\mu_i)_i$ based on their distance to the camera $d_i$.
% This list is then also populated with the samples from the NeRF $f_\theta$.
% Following this, we apply the same alpha compositing method as outlined in~\cref{alpha}.

To enforce disentanglement between the object and the human, we render the NeRF $f_\theta$ separately to obtain a human-only view $\bm x_{\text{H}}=g_{\text{H}}(\theta)$.
We then apply a guidance on $\bm x_\text{H}$ to ensure it contains a \emph{single} \emph{complete} human body, preventing the NeRF from generating extra limbs or additional people that might be occluded by the object in the full scene rendering $\bm x_\text{HO}$ (see~\cref{sec:ablations} for its effect).

% We also model the background using an MLP with learning rate scaled down by $10\times$ as in Magic3D~\cite{lin2023magic3dhighresolutiontextto3dcontent}.

% The colors form a rendered image $\bm x_{\text{L}}=g_{\text{L}}(\theta)$ that consists of the human model given by the NeRF.

% To render the global scene $\bm x_{\text{G}}=g_{\text{G}}(\theta)$ that consists of the human model and the object mesh $M_{\text{Obj}}$, we determine if each ray $R$ intersects $M_{\text{Obj}}$ at a point $\bm\mu_I$, and, if so, calculate $d_I=\Vert\bm\mu_{I}-\bm\mu_c\Vert$ using traditional rendering methods.
% We then obtain the RGBA value $\bm c_I,\tau_I$ of $M_{\text{Obj}}$ at point $\bm\mu_I$ and insert them to the list $(\bm\mu_i,\bm c_i,\tau_i)_i$ by inserting $d_I$ to the sorted distances $(d_i)_i$. We hence differentiably obtain the rendered image $\bm x_{\text{G}}$ with both the NeRF and the mesh $M_{\text{Obj}}$.

% We don't differentiate time steps between MVDream and IF
\subsection{Guidance Mixture}
\label{sec:guidance-mixture}
Multi-view text-to-image diffusion models are superior in generating 3D assets~\cite{shi23mvdream, zheng2023free3D, li2024instant3d, melas-kyriazi2024IM3d, voleti2024sv3d, gao2024cat3d} when compared to their single-view counterparts~\cite{poole2022dreamfusion, wang2023score, melas2023realfusion}.
Yet due to extensive fine-tuning on object-centric renderings, these multi-view models often struggle to capture the intended interaction $r$ between humans and objects.
Conversely, state-of-the-art single-view diffusion models~\cite{deepfloyd2023if,esser2024stablediffusion3} exhibit a more fine-grained understanding of the conditioning prompts, including the textual description of the interaction.
% We have empirically observed that this also applies to the understanding of the semantic relation $r$.
However, relying solely on them leads to 3D inconsistencies, including the multi-face Janus problem, where a person is depicted with multiple faces to accommodate different viewing angles.
To address this, we integrate the SDS guidance from both a multi-view model, MVDream~\cite{shi23mvdream}, and a larger single-view model, DeepFloyd IF~\cite{deepfloyd2023if}.

Specifically, we apply the SDS guidance using DeepFloyd IF on the scene rendering $\bm x_{\text{HO}}$ using the prompt $y_{\text{HO}}={}$``a photo of a person \{$r$\} a \{$M_{\text{Obj}}$\}, high detail, photography'', where \{$r$\} is the intended interaction (\eg, ``sitting on'') and \{$M_{\text{Obj}}$\} is the category of the object mesh (\eg, ``ball'').
We denote this loss as $\mathcal{L}_{\text{SDS-HO}}$.
For the human-only rendering $\bm x_{\text{H}}$, we employ a blend of SDS guidance using both DeepFloyd IF and MVDream, with the prompt $y_{\text{HO}}={}$``a photo of a person, high detail, photography''.
These losses are denoted as $\mathcal{L}_{\text{SDS-H}}$ and $\mathcal{L}_{\text{SDS-H-MV}}$, respectively.
We show that incorporating the MVDream guidance significantly enhances generation quality in~\cref{sec:ablations}.
Note that we avoid conditioning on the interaction or the object for $\mathcal{L}_{\text{SDS-H}}$ and $\mathcal{L}_{\text{SDS-H-MV}}$ to ensure that the NeRF represents only a \emph{single} person, as introducing additional objects can reduce the accuracy of the pose estimator when converting to the explicit mesh representation.

\subsection{Regularizers} \label{sec:regularizers}
To further encourage realistic HOI generations, we introduce two regularizers.
We define the \emph{sparsity above threshold regularizer}:
\begin{equation}
    \mathcal{R}_{\text{SA}} \coloneqq \text{softplus}(\bar{\bm x}_{\text{H}} - \eta),
\end{equation}
where $\bar{\bm x}_{\text{H}}$ is the average opacity of the human-only rendering $\bm x_{\text{H}}$.
This regularizer controls the ``size'' of the human to ensure it occupies no more than $\eta \coloneqq 20\%$ of the camera field of view. 
The purpose of $\mathcal{R}_{\text{SA}}$ is to compel the NeRF $f_\theta$ to contain only the human; without $\mathcal{R}_{\text{SA}}$, we observe that the NeRF tends to expand and cover the object mesh $M_{\text{Obj}}$ (see \ref{sec:ablations}).
To ensure robustness, we adjust the camera distance based on its randomly sampled focal length, ensuring that the 3D unit cube consistently occupies the same area in the renderings regardless of the focal length.

The \emph{intersection regularizer} $\mathcal{R}_{\text{I}}$ computes the average density $\tau$ (as predicted by $f_\theta$) of all ray points $\bm\mu$ \emph{inside} the object mesh $M_{\text{Obj}}$. Informally, it measures the volume of intersection between the human NeRF and the object. $\mathcal{R}_{\text{I}}$ discourages the model from generating body parts or other objects inside $M_{\text{Obj}}$, which would be invisible in $\bm{x}_{\text{HO}}$.

\subsection{Iterative Pose Optimization}
\label{sec:optimization}
Our objective function is a weighted sum of SDS guidances and regularizers:
\begin{multline}
\mathcal{L}(\theta) =
\lambda_{\text{SDS-HO}}\mathcal{L}_{\text{SDS-HO}}(g_{\text{HO}}(\theta))
+
\lambda_{\text{SDS-H}}\mathcal{L}_{\text{SDS-H}}(g_{\text{H}}(\theta))
\\+
\lambda_{\text{SDS-H-MV}}\mathcal{L}_{\text{SDS-H-MV}}(g_{\text{H}}(\theta))
\\+
\lambda_{\text{SA}}\mathcal{R}_{\text{SA}}(g_{\text{H}}(\theta))
+
\lambda_{\text{I}}\mathcal{R}_{\text{I}}(\theta).
\label{eq:loss}
\end{multline}
We initialize the NeRF $f_{\theta_0}$ with a density bias centered at the origin.
Throughout the optimization process, the implicit NeRF is periodically converted (every $10$k optimization steps) into its explicit form as a posed human mesh $M_{\xi_t}$ for identity grounding.
The NeRF is then re-initialized from $M_{\xi_t}$ and further refined by minimizing the loss $\mathcal{L}(\theta)$.

% Initially, the NeRF representation $f_{\theta_0}$ of the human is initialized with a density bias toward the origin, forming a ball.
% This is placed in a scene with the object mesh $M_{\text{Obj}}$. 
% Subsequently, it is optimized by minimizing the objective function $\mathcal L$, as defined in \cref{eq:loss}, for 10,000 steps. 
% The human NeRF $f_{\theta_0}$ is then converted into its explicit representation as a posed mesh $M_{\xi_0}$ by estimating its parameters $\xi_0$, and subsequently converted back into an implicit NeRF $f_{\theta_1}$, as described in \cref{sec:implicit-explicit}.
% The updated NeRF $f_{\theta_1}$ is further refined using SDS for an additional 10,000 steps.

This iterative conversion between implicit and explicit representations is repeated for $T$ times throughout the optimization.
After the first NeRF re-initialization, we disregard the human-only SDS terms $\mathcal{L}_{\text{SDS-H}}$ and $\mathcal{L}_{\text{SDS-H-MV}}$ from the loss $\mathcal{L}(\theta)$ since the human is already well-positioned, significantly reducing computational overhead.
We also decrease the noise level and render at a higher resolution (see \cref{sec:impl-details}).
The final output is the posed human mesh $M_{\xi_T}$.

% we consider only $\mathcal{L}_{\text{SDS-HO}}$ and the regularizers $\mathcal{R}_{\text{SA}}$ and $\mathcal{R}_{\text{I}}$, while disregarding $\mathcal{L}_{\text{SDS-H}}$ and $\mathcal{L}_{\text{SDS-H-MV}}$ since the human is already well-positioned, which reduces computational overhead.
% In addition, we decrease the noise level and render high-resolution images.
% With a user-given skinned human identity (texture and shape parameters), we hence also output the posed human mesh $M_{\xi_T}$.

% namely SMPL+H parameters $\xi_0$, and then converted back to an implicit NeRF $f_{\theta_1}$ which is re-initialized to fit the body pose $M_{\xi_0}$, using methods described in~\cref{sec:implicit-explicit}.
% The new NeRF $f_{\theta_1}$ is further refined using SDS for 10,000 steps.

% This back-and-forth process is repeated for a few iterations (or until the user finds a satisfactory result).
% After the first NeRF re-initialization, we only consider $\mathcal{L}_{\text{SDS-HO}}$ and the regularizers $\mathcal{R}_{\text{SA}}$, $\mathcal{R}_{\text{I}}$ and disregard $\mathcal{L}_{\text{SDS-H}}$, $\mathcal{L}_{\text{SDS-H-MV}}$ (since the human is already in a good position), which reduces the computation budget (since we only render 1 view instead of 6 and avoid loading MVDream).
% We also reduce the time-step and increase the rendering resolution after the NeRF re-initialization (see~\cref{sec:impl-details}).

\section{Experiments}
\label{sec:experiments}
We conduct experiments on the subject-driven generation of 3D human-object interactions task. We use the SMPL+H model~\cite{loper15smpl, romero2017mano} as the underlying rigged human model. We evaluate our proposed pipeline against several baselines and ablation settings.

% Each run in total requires about 6--8 hours to train on an NVIDIA A40 or A6000 GPU, with 5 hours for the NeRF training stage, and 1 hour for each iteration of the human model fitting and DMTet training stage.

\subsection{Dataset}
We select meshes suitable for human interaction from Sketchfab~\cite{Sketchfab}, a website that hosts CC-licensed 3D models uploaded by the community.
We exclude meshes uploaded \emph{before} the training cutoff date of DeepFloyd IF and MVDream to eliminate data contamination. We then use GPT-4~\cite{openai23gpt4} to generate a set of prompts for the chosen meshes corresponding to a variety of human-object interactions, and filter out prompts that are too similar or impossible (\eg, ``juggling balls'' as there is only a single ball mesh), yielding a set of $12$ prompts for comparisons and ablation studies.
We manually position and scale $M_{\text{Obj}}$ for each prompt such that it is natural to generate a human interacting with $M_{\text{Obj}}$. We source a set of skinned human models as input to our pipeline from SMPLitex~\cite{casas2023smplitex}.

\subsection{Implementation Details}
\label{sec:impl-details}
We estimate the SMPL+H~\cite{romero2017mano} pose parameters $\xi$ from NeRF $f_\theta$ by using rendered images $\bm x_i$ from different camera angles $c_i$ as in SMPLify-X~\cite{pavlakos2019smplx}.
This is done by optimizing the parameters $\xi$ such that the resulting keypoints, when projected to 2D space by $c_i$, match with keypoints predicted from $\bm x_i$ using OpenPose~\cite{cao2017_openpose,wei2016_openpose_cpm,simon2017_openpose_hand}.
Our implementation is adapted from an extension~\cite{zheng2020pamir} of SMPLify-X that can train on multiple images.

% \subsection{Hyperparameters} \label{sec:hyperparameters}
We set $\lambda_{\text{SDS-HO}}=0.9$, $\lambda_{\text{SDS-H}}=0.05$, and $\lambda_{\text{SDS-H-MV}}=0.05$. 
For the first stage (\ie, before any implicit-explicit conversion and NeRF re-initialization),
we set 
$\lambda_{\text{I}}=1$ and $\lambda_{\text{SA}}=10000$ to enforce a strict volume of the NeRF.
After the NeRF re-initialization,
$\lambda_{\text{SA}}$ is decreased to $1000$ for more flexibility.
The SDS noise level (\ie, timestep) is sampled from $\mathcal{U}(0.02, 0.98)$ during the first stage and $\mathcal{U}(0.02, 0.70)$ subsequently.
These hyper-parameters apply to all $12$ prompts and we do \emph{not} tune them individually for each prompt.

During the first stage, we render $64\times64$ images for the former half of the training and $256\times256$ images for the latter half, following MVDream~\cite{shi23mvdream}. After NeRF re-initialization, the resolution is increased to $512\times512$ to achieve finer quality, as in Magic3D~\cite{lin2023magic3d}.
We do not use shading for NeRF~\cite{poole2022dreamfusion} as it is costly to compute.

\subsection{Results}
\begin{figure}[t]
\centering
\includegraphics[width=.99\linewidth]{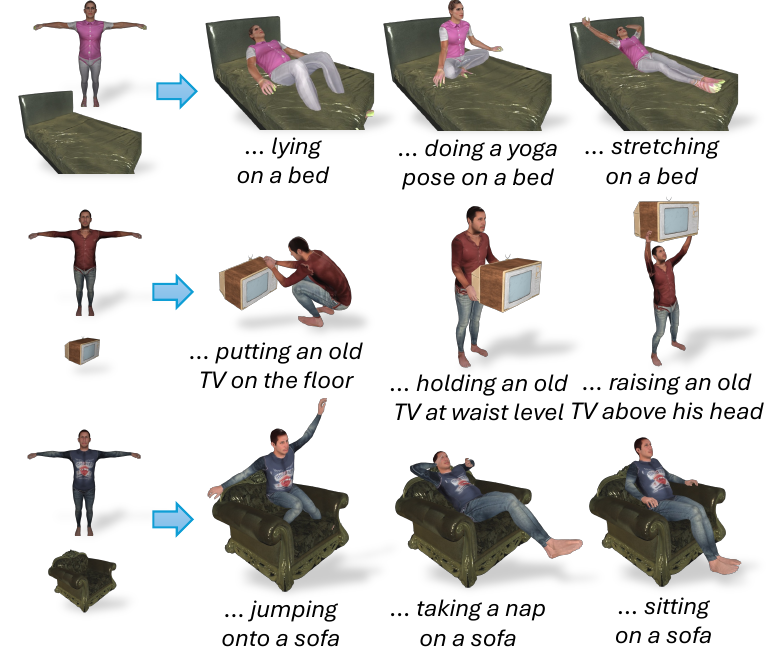}
\caption{\textbf{Additional Results.} We demonstrate \method's ability to control the pose based on different textual conditions.}
\label{fig:results}
\end{figure}

We present qualitative results in \cref{fig:teaser} and \cref{fig:results}.
We find \method{} can adapt to a wide variety of human-object interactions, including those that are complex or rare (\eg, ``doing push-ups with hands on a ball'').
It also automatically recognizes differences in object geometry (\eg a sports motorcycle \vs a cruiser, \cref{fig:teaser}) without mentioning the difference in the prompt (``riding a motorcycle'') and adapts the human pose accordingly.
Given the same object, it can also respond to differences in the prompts (``lying'' \vs ``stretching'' on a bed, \cref{fig:results}).
The open-world nature of these results contrasts with traditional methods for human-object interaction that only work on fixed categories of objects and a limited number of interactions as input.

\subsection{Comparisons} \label{sec:comparisons}

\begin{figure}[t]
\centering
\includegraphics[width=\linewidth]{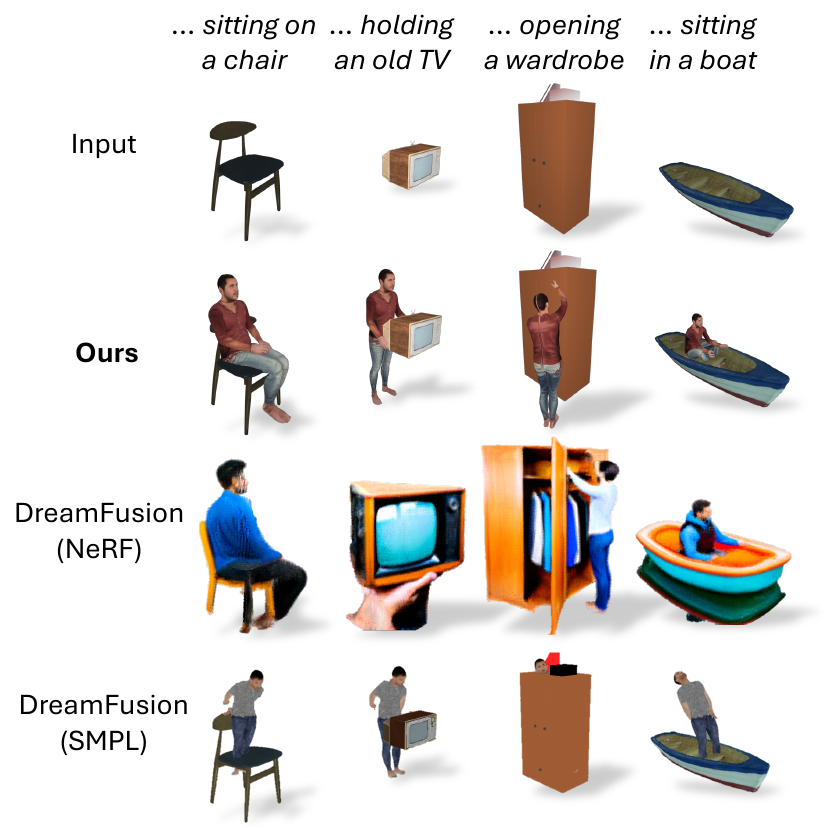}
\caption{\textbf{Comparison} with baselines: (\textbf{third row}) using DreamFusion to optimize a NeRF with the given object mesh inserted; (\textbf{last row}) using DreamFusion to optimize the pose parameters of the skinned human mesh directly.
See \cref{sec:comparisons} for discussions.
}
\label{fig:comparisons}
\end{figure}

We compare \method{} with baseline open-world approaches.
%in~\cref{fig:comparisons} and \cref{tab:comparisons}.
In particular, we consider (1) using DreamFusion SDS loss~\cite{poole2022dreamfusion} to fit a NeRF, and (2) using DreamFusion to fit SMPL+H~\cite{loper15smpl, romero2017mano} pose parameters directly, as the baselines.
Both baselines use DeepFloyd IF~\cite{deepfloyd2023if} as the guidance model for SDS.

We show qualitative comparison in~\cref{fig:comparisons}.
We first note that DreamFusion (NeRF) does \emph{not} achieve our goal of \emph{subject-driven} human-object interaction generation, as it \emph{cannot} preserve the identity and structure of the human mesh.
It has limited success on some prompts (first and last columns) and also fails on others (second and third columns).
We find that DreamFusion frequently generates a NeRF that is overly large and overtakes the object mesh, and its outputs are not view-consistent (\eg, the three-faced TV).
% These problems and our methods to fix them are expanded in ablations (\cref{sec:ablations}).
DreamFusion (SMPL), on the other hand, completely fails to produce correct poses.
This occurs because the image-space gradients provided by SDS \emph{cannot} be effectively backpropagated into meaningful updates in the bone rotation space.
For example, when adjusting the position of an arm, SDS does not provide a gradient that gradually shifts the arm. Instead, it deletes the arm and re-generates it in a new position, which may be far from the original. As a result, no gradients from the re-generated arm propagate through differentiable mesh rendering to the bone rotation parameters.
% We discuss this further in~\cref{sec:intro}.

\begin{table}
\centering
\caption{\textbf{Quantitative Comparisons and Ablations.}
We compute the mean and standard deviation of CLIP similarities between the renderings of generated HOIs and the corresponding text prompts.
% addressed in sec 4.1 \ray{12 prompts seem to be inconsistent with the 37 we wrote in sec 4.1, might be trigger a red flag among reviewers.}
} \label{tab:ablations} \label{tab:comparisons}
\begin{tabular}{lcc}
\hline
Method &  CLIP Score $\uparrow$ \\ \hline
DreamFusion (NeRF) & 0.2915${}\pm{}$0.0182 \\
DreamFusion (SMPL) & 0.2511${}\pm{}$0.0247 \\
% \hline
\rowcolor{gray!20}
\textbf{\method{} (Ours)}           & \textbf{0.2932}${}\pm{}$0.0239 \\
% \hline
\quad Remove $\mathcal{R}_{\text{SA}}$, $\mathcal{R}_{\text{I}}$     & 0.2879${}\pm{}$0.0315                  \\
% \hline
\quad Remove $\mathcal{L}_{\text{SDS-H}}$, $\mathcal{L}_{\text{SDS-H-MV}}$ & 0.2869${}\pm{}$0.0306                  \\
% Remove NeRF re-fitting & $?$          \\
\hline
\end{tabular}
\end{table}

To compare quantitatively, we render our final scenes, which consist of the object mesh and the posed human mesh, from multiple viewpoints. 
We then measure the average CLIP similarity~\cite{radford2021learning,hessel2022clipscore} between our prompts and the renderings of the composed scene. 
Our method consistently outperforms the baselines as shown in \cref{tab:ablations}. 

\subsection{Ablations} \label{sec:ablations}
\begin{figure}[t]
\centering
\includegraphics[width=\linewidth]{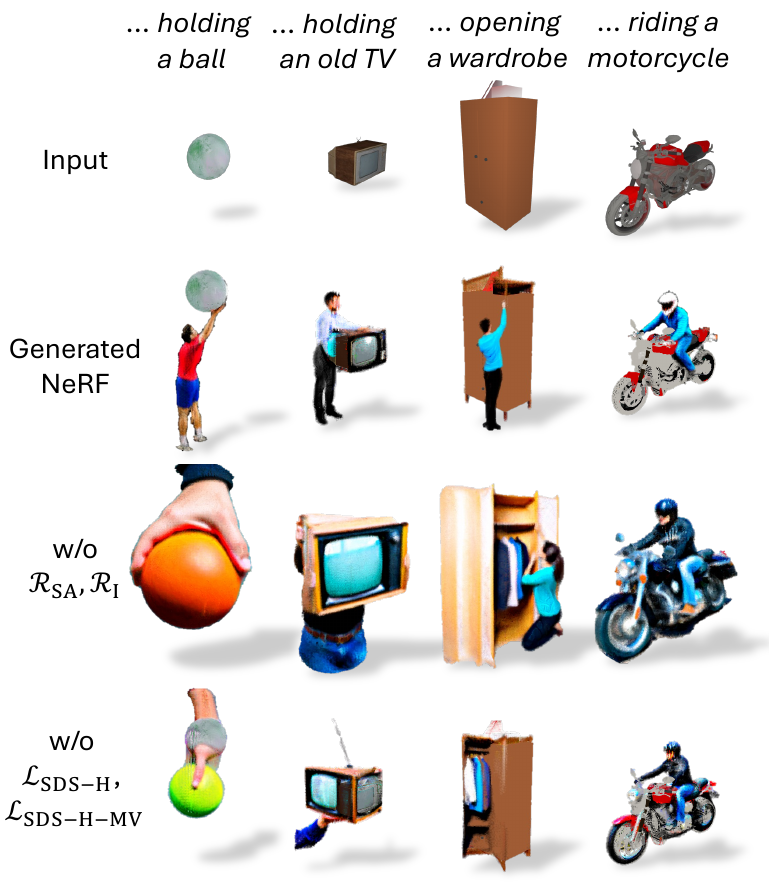}
\caption{\textbf{Ablations.} We visualize output of first-round NeRF optimization (\ie, $f_{\theta_0}$).
(\textbf{Row 2 \vs row 3}) Our regularizers (\cref{sec:regularizers}) effectively prevent the NeRF from encroaching upon and overriding the mesh.
(\textbf{Row 2 \vs row 4}) The human-only SDS motivates a complete human while MVDream enhances view consistency (note the TV and wardrobe in row 4 have multiple faces).}
\label{fig:ablations}
\end{figure}

To demonstrate the effectiveness of our techniques, we test our pipeline in the following ablation settings:
(1) without regularizers $\mathcal{R}_{\text{SA}}$, $\mathcal{R}_{\text{I}}$;
and (2) without the human-only rendering $g_H$ (\ie, removing $\mathcal{L}_{\text{SDS-H}}$, $\mathcal{L}_{\text{SDS-H-MV}}$).
We present qualitative results in~\cref{fig:ablations} and quantitative results in~\cref{tab:ablations}.
% (3) with only one stage of implicit optimization without re-fitting the NeRF.

% We find that the outputs of the NeRF implicit optimization are very different, as seen in~\cref{fig:ablations}.
The regularizers $\mathcal{R}_{\text{SA}}$, $\mathcal{R}_{\text{I}}$ effectively limit the size of the generated NeRF of the human; without them, the NeRF expands to include both the human and the object, covering and overriding the existing object mesh $M_{\text{Obj}}$, resulting in wrong human and object positions (\cref{fig:ablations} third row).
On the other hand, SDS on the human-only rendering $g_{\text{H}}$ ensures the presence of a complete human body, as evidenced by the absence of the person in the wardrobe example or the presence of only an arm in the ball and TV examples (\cref{fig:ablations} last row).
Furthermore, the MVDream~\cite{shi23mvdream} guidance enforces directional consistency, thereby eliminating the multi-face Janus problem, as seen from the three-faced TV and wardrobe in the last row.
MVDream takes camera positions as input, and since most objects in its training data (Objaverse~\cite{deitke22objaverse} renderings) face the forward direction, this also gives a canonical ``front'' direction for generating the HOI.
See Sup.\ Mat.\ for details.

\begin{figure}[t]
\centering
\includegraphics[width=.99\linewidth]{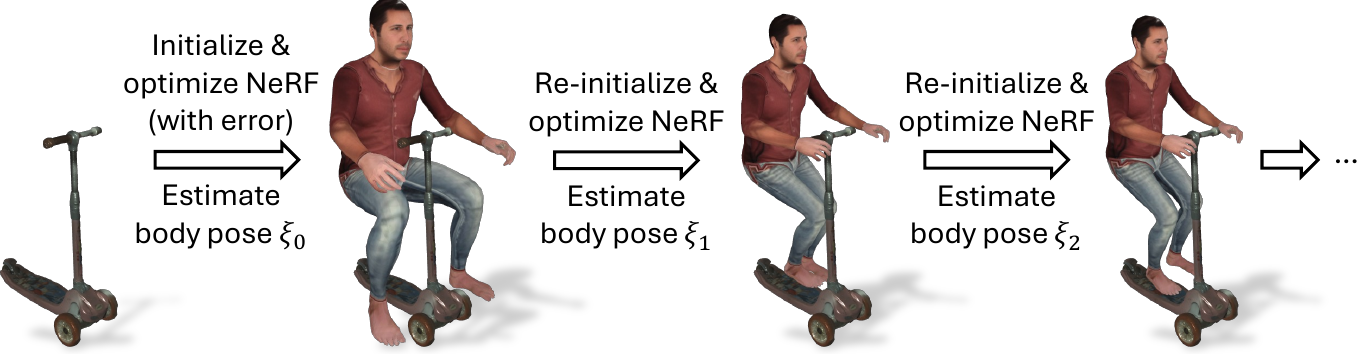}
\caption{
\textbf{Effect of Iterative Pose Optimization.}
When the initial human pose is sub-optimal,
by translating the explicit skinned mesh back to the implicit NeRF representation and optimizing the NeRF further, the pose improves significantly.
Note the movement of the feet toward the deck and the hands toward the handle. 
% \ray{I think we don't need to write the words 3 times.} -> I think it's clearer this way
}
\label{fig:stages}
\end{figure}

In~\cref{fig:stages}, we visualize the effect of iterative pose optimization (\cref{sec:optimization}) on a sub-optimal initial NeRF: the dual implicit-explicit optimization process improves the body pose.
The improvement is attributed to the capability of SDS to effectively manipulate shapes represented as NeRFs, together with the re-initialization of NeRF through the corresponding posed mesh, which grounds the NeRF.
 % eg the CLIP score for the scooter example {fig:stages} sometimes decreased even when the result becomes visibly better

\subsection{Limitations}\label{sec:limitations}
Similar to other 3D generation methods reliant on 2D guidance, the quality of our results is constrained by the ability of image diffusion models to accurately follow prompts. Future improvements in the underlying guidance models could lead to significant advancements in our approach.

\method{} depends on existing pose estimators (\ie, OpenPose~\cite{cao2017_openpose, simon2017_openpose_hand, wei2016_openpose_cpm}) to regress the pose from the implicit NeRF. Consequently, our results depend on their ability to predict keypoints in the NeRF renderings. It is also limited by the training data and targets (\ie, humans) of these estimators.
We leave it as future work to develop an automatic method that can find correspondences between two articulated shapes to perform this alignment. 
This would enable us to extend \method{} to other categories where pose estimators are not readily available, while also circumventing the challenges associated with non-human data collection.

\section{Conclusions}
\label{sec:conclusions}

We have proposed \method{}, an approach for posing rigged human models to interact naturally with given 3D objects conditioned on textual prompts.
At the core of \method{} is a dual implicit-explicit representation that enables the use of SDS to optimize the explicit pose parameters of skinned meshes, supplemented by a guidance mixture technique and novel regularization terms.
We experiment on a diverse set of prompts and demonstrate our method achieves superior generation quality compared to various baseline approaches.
Ablation studies verify the importance of the components that contributed to this improvement.
Our work has the potential to simplify the creation of virtual environments populated by realistically interacting humans, which can be used in many applications, such as film and game production.
\paragraph{Acknowledgments.}
The authors would like to thank Lorenza Prospero for her helpful feedback on the manuscript.

{
    \small
    \bibliographystyle{ieeenat_fullname}
    \bibliography{main}
}

\clearpage
\appendix
\setcounter{page}{1}
% \maketitlesupplementary
% replaced \maketitlesupplementary with:
\twocolumn[
\centering
\Large
\textbf{\thetitle}\\
\vspace{0.5em}Supplementary Material \\
\vspace{1.0em}
% \twocolumn[
{
    \centering
    \includegraphics[width=0.99\textwidth]{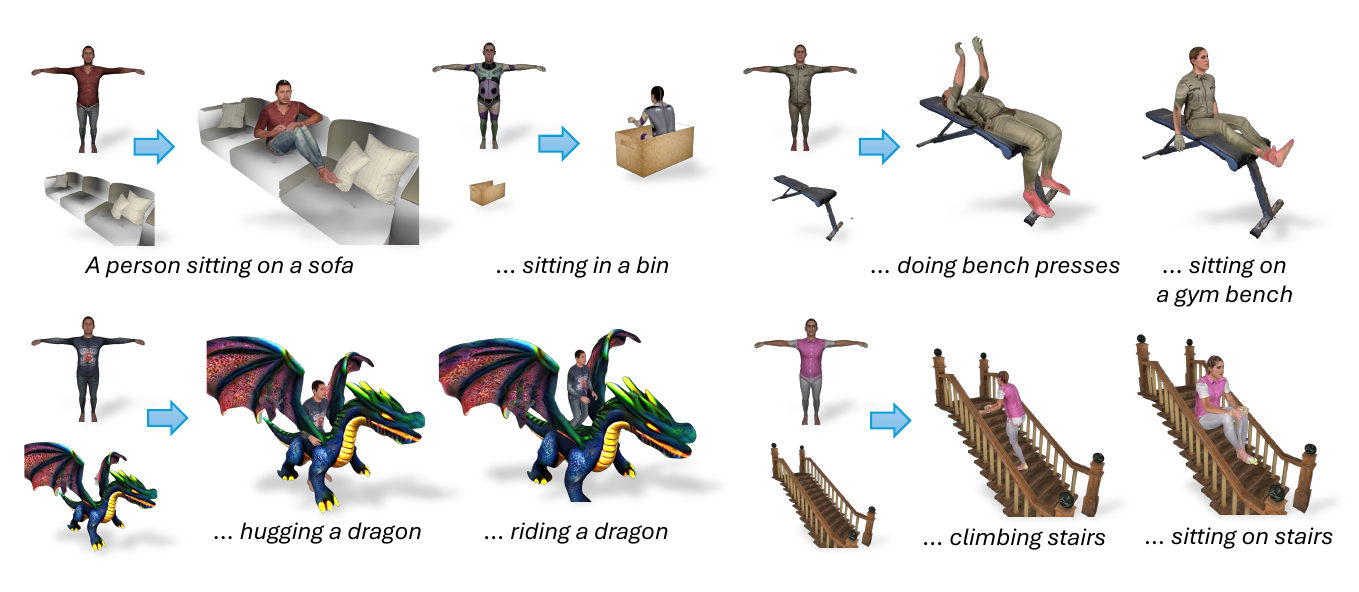}
    \captionof{figure}{
    \textbf{Additional Results.}
    \method~is able to generate HOIs for diverse objects and corresponding prompts.
    }
    \label{fig:additional-results}
}
% ]

% \begin{figure*}
%     \centering
%     \includegraphics[width=\linewidth]{figures/supmat/add-results.pdf}
%     \caption{
%     \textbf{Additional results by \method.} We demonstrate our pipeline's ability to adapt to diverse prompts and objects
%     % \tom{this is a great figure and should be on the first page of supmat} 
%     }
%     \label{fig:additional-results}
% \end{figure*}

\bigbreak
]

\section{Additional Results}
\label{sec:supmat-additional-results}

\paragraph{Additional qualitative results.}
Additional results on a variety of prompts and objects can be found in~\cref{fig:additional-results}.
\method{} is capable of realistically deforming the human pose to interact with these objects faithfully to the corresponding textual prompts.

\paragraph{Failure cases.}

\begin{figure}[t]
\centering
\includegraphics[width=\linewidth]{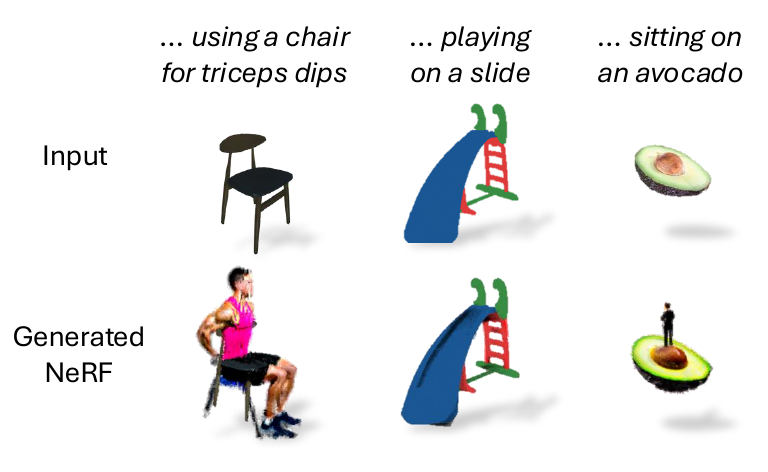}
\caption{\textbf{Failure Cases.}
We visualize failed outputs from the first-round NeRF optimization (\ie, $f_{\theta_0}$).
Our pipeline is unlikely to recover if the NeRF from the initial round fails to capture the approximate spatial relationship between the human and the object.
}
\label{fig:failure-analysis}
\end{figure}

We show some cases where \method{} failed in~\cref{fig:failure-analysis}. From a manual inspection, in most cases this was due to the underlying diffusion model not understanding the semantic composition, because it was too complex, vague, or exotic (respectively, in~\cref{fig:failure-analysis}). In other cases this was due to the pose prediction (SMPLify-X~\cite{pavlakos2019smplx}) not working properly. Therefore we believe an improvement to either would improve \method{}'s ability to generate realistic HOIs.

\paragraph{Additional comparisons.}
\label{sec:supmat-additional-comparison}

\begin{figure}[t]
\centering
\includegraphics[width=\linewidth]{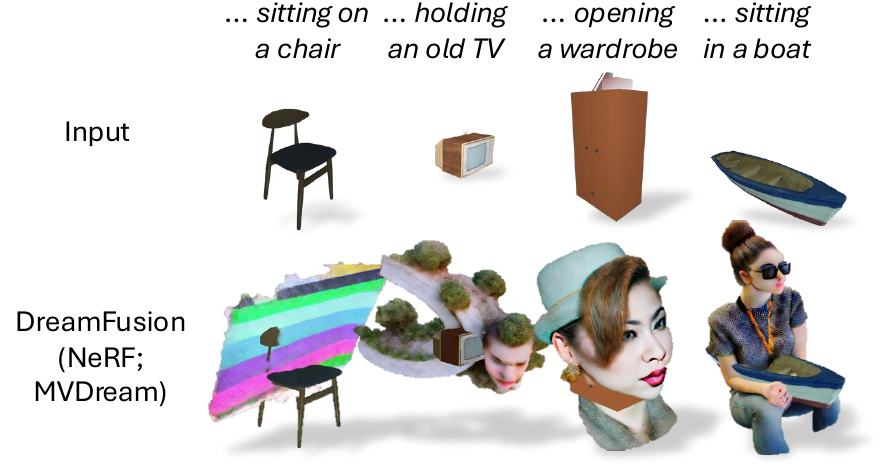}
\caption{
\textbf{Additional Comparison.}
Following~\cref{fig:comparisons}, we additionally compare \method~to a DreamFusion baseline where we use only MVDream~\cite{shi23mvdream} for the diffusion guidance without DeepFloyd IF~\cite{deepfloyd2023if}.}
\label{fig:additional-comparisons}
\end{figure}

In our baseline comparisons (\cref{sec:comparisons}), we considered comparing to the case where we generate a NeRF by DreamFusion using DeepFloyd IF guidance. We showed that, although mostly related to the prompt, the outputs often had problems like not view-consistent or being too large. We now compare against the same baseline but with IF replaced with MVDream guidance, in~\cref{fig:additional-comparisons}.
Note that MVDream guidance is able to produce very detailed and view-consistent NeRFs, but fails to understand the basic compositional relations in the prompt (\eg, ``sit''). This provides motivation to use DeepFloyd IF as our base model for better textual understanding.

\section{Discussions}
\label{sec:supmat-discussions}
\paragraph{Failure analysis of direct optimization.}

\begin{figure}[t]
\centering
\includegraphics[width=\linewidth]{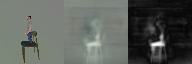}
\caption{
\textbf{Visualization} of SDS gradient. Left: rendered $\bm{x}_{\text{HO}}$ ($M_\xi$ with $M_{\text{Obj}}$);
Middle: gradient $\nabla_{\bm{x}}\mathcal{L}_{\text{SDS-HO}}$;
Right: norm of the gradient.}
\label{fig:sds-grad}
\end{figure}

The most straightforward way to solve the HOI generation task we propose is to only use explicit SMPL (or any other body model) pose parameters $\xi$ instead of an implicit NeRF $f_\theta$ as the object for optimization. In this solution, we render the resulting SMPL mesh $M_\xi$ with an object mesh $M_{\text{obj}}$, and use SDS to directly optimize $\xi$. This avoids the problem of translation between explicit and implicit forms and makes optimization much faster (\eg SMPL only has 69 pose parameters~\cite{loper15smpl}).

However, in our extensive tests, this method does not work at all (even if we initialize from a near-optimal pose, $\xi$ regresses to a nonsensical pose as in~\cref{fig:comparisons}; additional changes like making vertex colors and global position learnable do not help either). This was observed in~\cref{sec:comparisons}.

To illustrate the reason, we monitor the SDS loss and gradient during the optimization of $\xi$ in~\cref{fig:sds-grad}, for the prompt ``a person sitting on a chair''. In the middle and right panels, the guidance tries to add legs to the position where legs would be expected, on the seat and to its front. However, there is no way for this gradient to add legs to be propagated to $\xi$, because $\xi$ can only receive gradients on pixels that the rendered $M_{\xi}$ occupies. In other words, the tendency of diffusion models to add and delete limbs globally, instead of gradually moving a limb, means that SDS is not suitable for optimizing $\xi$ directly. This necessitates the dual implicit-explicit optimization we propose.

\paragraph{MVDream front direction.}

\begin{figure}[t]
\centering
\includegraphics[width=\linewidth]{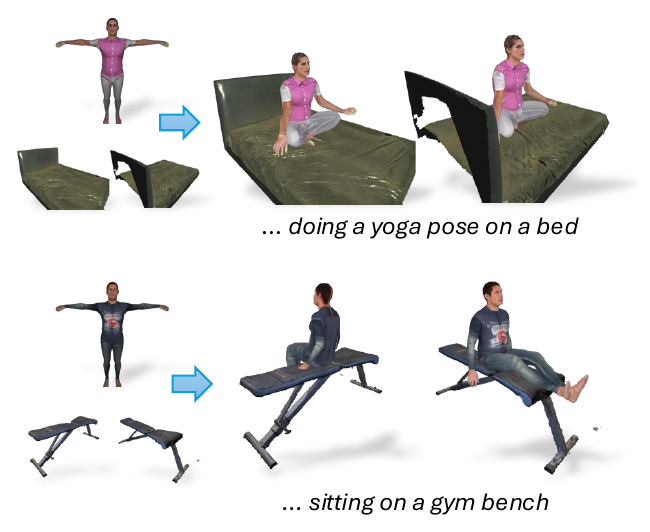}
\caption{
\textbf{Front Direction Control.}
MVDream~\cite{shi23mvdream} guidance enables us to give an implicit ``front'' direction: the generated humans consistently face the $\pm x$ direction regardless of the orientation of the object.}
\label{fig:front-direction}
\end{figure}

We claimed in~\cref{sec:ablations} that MVDream takes camera positions as input and based on bias in its training data, MVDream gives a prior of a ``front'' direction (in $+x$ direction) for generating the HOI. We demonstrate this in~\cref{fig:front-direction}. This gives us the ability to control the forward face of the entire HOI (typically the direction the person is facing, or its opposite) with respect to the object by rotating the mesh $M_{\text{Obj}}$ in the generation.

\section{Technical Details}
\label{sec:supmat-technical-details}
% done \tom{technical details should go last as they are the least interesting}

\paragraph{Optimization.} We follow MVDream~\cite{shi23mvdream} and optimize the initial NeRF in $2$ stages.
In each stage, we optimize the NeRF with AdamW optimizer (learning rate and weight decay are both set to $0.01$) for $5000$ steps.
We render $64\times 64$ and $256\times 256$ images and use batch size $8$ and $4$ respectively in two stages.
After NeRF re-initialization, MVDream guidance is no longer used, and we increase the rendering resolution to $512\times 512$, reduce the batch size to $1$, and decrease the learning rate to $0.001$.

The field of view $f$ of the camera in each optimization step is sampled uniformly at random within $[15^\circ, 60^\circ]$, and the camera distance to origin is set to $D/\tan(f/2)$ where the denominator is such that a unit volume in the 3D space corresponds roughly to a fixed area in the 2D space, as in~\cref{sec:regularizers} for $\mathcal{R}_{\text{SA}}$ to work properly, and  $D\sim[0.8,1.0]$ is a perturbation. The elevation angle is sampled uniformly from $[0^\circ,30^\circ]$. Although not done in this work, we recommend lowering it to $[-30^\circ,30^\circ]$ if parts of the human may be below the object for better supervision.
For rendering views $\bm{x}_i$ of the NeRF for pose estimation (\cref{sec:impl-details}), we use an array of cameras with distance 3 to the origin, elevated at $40^\circ$.

The NeRF representing the human is constrained in a ball of radius 1. 
We initialize it to be at the origin. The number of parameters of the NeRF MLP ($f_\theta$) is about 12.6 million.

The background color is learned, with a lower learning rate $0.001$, and is replaced during training with a random color with probability 0.5 (increased to 1 after re-initialization) in training for augmentation.

The NeRF renderer uses one ray per 2D pixel and 512 samples per ray.

\paragraph{Guidance.}
For SDS, 
we use classifier-free guidance~\cite{ho2022cfg} with guidance weight set to $\omega=50$.
We include the \emph{negative} prompt ``missing limbs, missing legs, missing arms'' during optimization.

\end{document}